\documentclass[letterpaper, 10pt, conference]{ieeeconf}      
\usepackage[utf8]{inputenc}
\IEEEoverridecommandlockouts                              

\usepackage[export]{adjustbox}
\usepackage{graphicx}
\usepackage{mathtools}
\usepackage{xfrac}
\usepackage[english]{babel}
\usepackage[T1]{fontenc}
\usepackage{amsmath,amssymb,bbm} 
\usepackage{array}
\usepackage{mathrsfs}
\usepackage{cite}
\usepackage{multirow}
\usepackage{nccmath}
\usepackage{amsmath}
\usepackage{subfig}
\usepackage{multirow}
\usepackage{color}
\usepackage{url}
\usepackage[table]{xcolor}
\usepackage[ruled,vlined]{algorithm2e}
\newcommand{\squeezeup}{\vspace{-2.5mm}}

\SetKwInput{KwDefinition}{Definition}
\SetKwInput{KwInput}{Input}
\SetKwInput{KwOutput}{Output}
\SetKwBlock{with}{with}{end}

\title{\LARGE \bf A Multimodal Target-Source Classifier with Attention Branches\\ to Understand Ambiguous Instructions for Fetching Daily Objects
}

\author{Aly Magassouba, Komei Sugiura and Hisashi Kawai
 \thanks{National Institute of Information and Communications Technology, 3-5 Hikaridai, Seika, Soraku, Kyoto 619-0289, Japan  {\tt\small name.surname@nict.go.jp}} 
 } 
 
\newcommand*\Update{\color{black}}
\newcommand*\Done{\color{black}}


\begin{document}

\maketitle
\graphicspath{{figures/}} 
\begin{abstract}
In this study,  we focus on  multimodal language understanding for fetching instructions in the domestic service robots context. This task consists of predicting a target object, as instructed by the user, given an image and an unstructured sentence, such as ``Bring me the yellow box (from the wooden cabinet).'' This is challenging because of the ambiguity of natural language, {i.e.}, the relevant information may be missing or there might be several candidates. To solve such a task, we propose the multimodal target-source classifier model with attention branches (MTCM-AB),  which is an extension of the MTCM \cite{magassouba2019understanding}.  Our methodology uses the attention branch network (ABN) \cite{Fukui_2019_CVPR} to develop a multimodal attention mechanism based on linguistic and visual inputs.  Experimental validation using a standard dataset showed that the MTCM-AB outperformed both state-of-the-art methods and the MTCM. In particular, the MTCM-AB accuracy was 90.1\% on average while human performance was 90.3\% on the PFN-PIC dataset. 
\end{abstract}

\section{Introduction}
The current rise of life expectancy has increasingly emphasized the need for daily care and support.  Robots that can physically assist people with disabilities \cite{brose2010role} offer an alternative to overcoming the shortage of home care workers. This context has boosted the need for standardized domestic service robots (DSRs) that can provide necessary support functions as shown by \cite{piyathilaka2015human, smarr2014domestic,iocchi2015robocup}. 

However, one of the main limitations of DSRs is their inability to naturally interact through language. Specifically, most DSRs do not allow users to instruct them with various expressions relating to an object for fetching tasks. By overcoming this limitation, a user-friendly way to interact with DSRs could be provided to non-expert users.

Thus, our work focuses on multimodal language understanding for fetching instructions (MLU-FI). This task consists of predicting a target  instructed in natural language, such as ``{\it Bring me the yellow box from the wooden cabinet}.''  \Update MLU-FI is challenging because of the ambiguity of natural instructions, that is, relevant information may be missing, implicit or simply paraphrased. Using free-form language naturally induces ambiguity because of the many-to-many mapping between the linguistic and physical world which makes it difficult to accurately infer the user's intention.
\Done

\begin{figure}[tp]
   \centering
      \includegraphics[width=0.9\columnwidth]{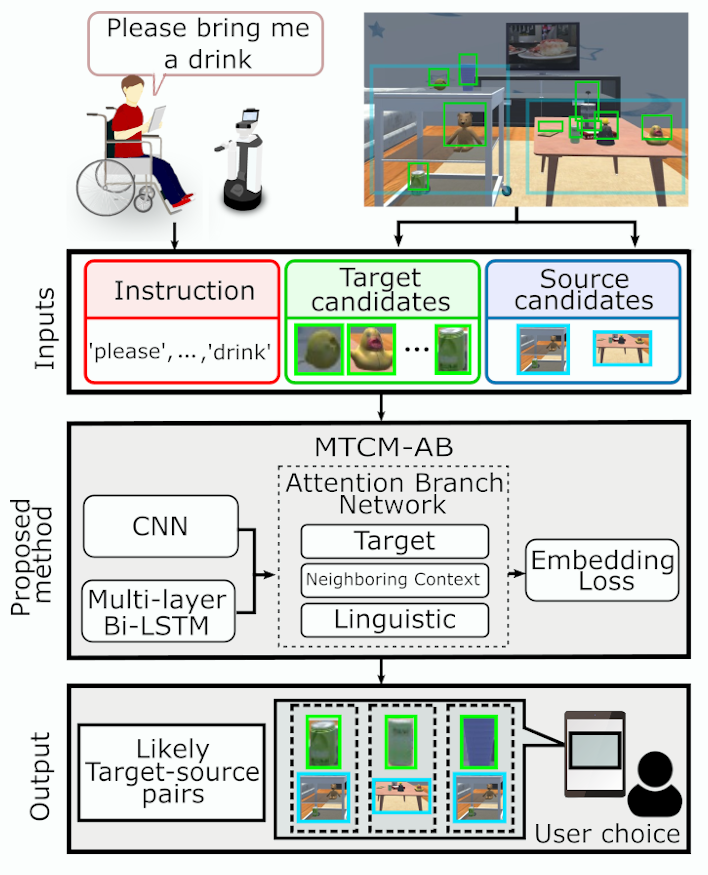}
      \caption{\small MTCM-AB overview: the attention branch network is used to improve fetching instruction comprehension}
   \label{fig:architecture}
 \squeezeup
\squeezeup
 \end{figure}

In this paper, we propose the multimodal target-source classifier model with attention branch (MTCM-AB) which is an extension of the MTCM proposed in \cite{magassouba2019understanding}. Our methodology uses the attention branch network (ABN) proposed in \cite{Fukui_2019_CVPR}. The ABN is an image classifier, inspired by class activation mapping (CAM) \cite{zhou2016cvpr} structures, that generates attention maps. The ABN is composed of an attention branch that infers an attention map and a perception branch that classifies images. We extended this architecture with the MTCM-AB where several visual and linguistic attention branches are proposed to infer visual and linguistic attention maps. From these multimodal attention maps, the model accuracy is enhanced and can be visualized through the attended areas of the visual input. A video is available at this URL\footnote{\protect\url{https://youtu.be/AMwqOEmDUjE}}.

The main contributions of this paper are summarized as follows:
\begin{itemize}
 \item We propose the MTCM-AB which extends the MTCM with the ABN. 
\item We introduce a Multimodal ABN architecture that combines both linguistic and visual attention mechanisms for MLU-FI.
\item We propose a visual explanation from the MTCM-AB given the input sentence and visual scene.
\end{itemize}

\section{Related Work}
There have been many attempts in the field of robotics focused on grounded communication with robots (e.g., \cite{magassouba2018multimodal, cohen2019grounding}). Grounding a user's intention requires linguistic inputs but also proprioceptive senses (e.g., vision) and contextual knowledge.  



Similarly to many studies, we are interested in understanding fetching instructions in everyday environments. Recent studies have addressed multimodal language understanding (MLU) by using visual semantic embedding for visual grounding \cite{nagaraja2016modeling, yu2017joint, hatori2018interactively, Shridhar-RSS-18, magassouba2019understanding}, visual question answering\cite{antol2015vqa} or caption generation \cite{vinyals2015show}. This approach embeds the visual and linguistic features into a common latent space. In \cite{nagaraja2016modeling} the authors proposed a long short-term memory (LSTM) network  to learn the probability of a referring expression, while a unified framework for referring expression generation and comprehension was introduced in \cite{yu2017joint}.
Inter and intra self-attention mechanisms are explored in \cite{yu2019multimodal} for referring expression comprehension.
In  robotics,  the authors of \cite{hatori2018interactively} developed a target prediction method from natural language in a pick-and-place task environment, with additional dialogue. Similarly \cite{Shridhar-RSS-18} tackled the same kind of problem using a two-stage model to predict the likely target from the language expression and the pairwise relationships between different target candidates. In our previous work \cite{magassouba2019understanding}, we proposed to use both the target and source candidates to predict the likely target in a supervised manner.

The MTCM-AB is inspired by the ABN \cite{Fukui_2019_CVPR}. The ABN is based on class activation mapping (CAM) networks \cite{zhou2016cvpr, Selvaraju_2017_ICCV}. This line of research focuses on the production of image masks that, overlaid onto an image, highlight the most salient portions with respect to some given query or task. In essence, the CAM purpose is to identify salient regions of a given label in an image classifier for visual explanation. The ABN builds visual attention maps from this approach.

Attention mechanisms have also been used differently in image and natural language processing (NLP). In the context of image captioning, the authors of \cite{xu2015show} generated image captions with hard and soft visual attention. This approach learns the alignment between the salient area of an image and the generated sequence of words. Multiple visual attention networks were also proposed in \cite{Yang_2016_CVPR} for solving visual question answering. However, most of these approaches use only a single modality for attention: visual attention. By contrast, recent studies in multimodal language understanding have shown that both linguistic and visual attention are beneficial for question-answering task \cite{nguyen2018improved, lei2019tvqa+} or visual grounding \cite{akbari2019multi, yu2018mattnet}. Similarly  in \cite{hori2017early}, an attention method that performs a weighted average of linguistic and image inputs is introduced. Against this context, a multimodal attention branch network has been proposed in \cite{magassouba2019multi} for sentence generation. Unlike the former, the current study focuses on MLU-FI and adopts a different structure that is detailed in the following sections.

\section{Problem Statement}\label{sec:prob}

\begin{figure}[tp]
  \centering
   \subfloat[Bring me the bottle in the middle of the grey wagon]{\includegraphics[width=4. cm, height=3.5cm]{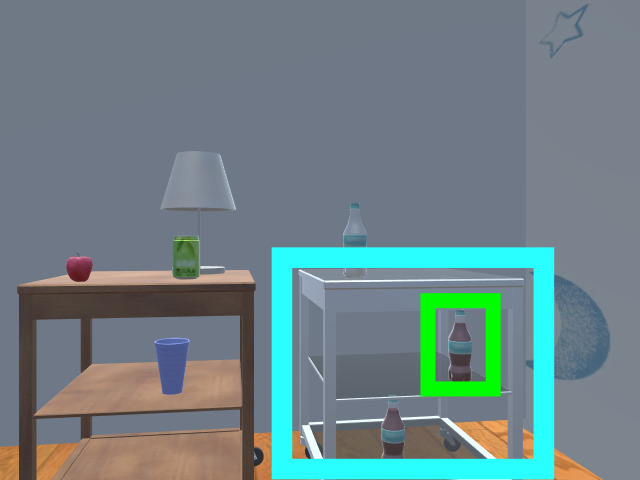}}\enskip
    \subfloat[Grab the sauce bottle from  the upper right box and place it...]{\includegraphics[width=4. cm, height=3.5cm]{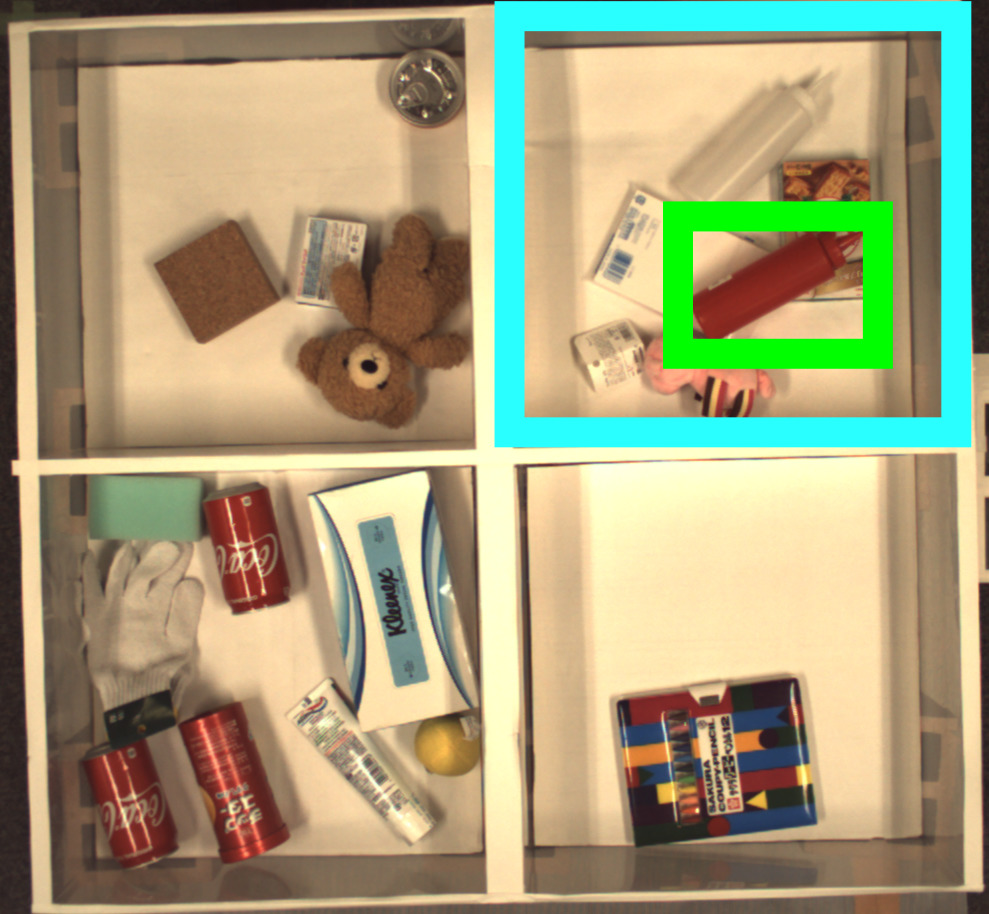}}
    \caption{\small  Samples of the WRS-PV (left) and PFN-PIC datasets (right) where the source and target are given.} \label{fig:sigverse} 
    \squeezeup
\squeezeup
\squeezeup
\end{figure}

The purpose of this study is understanding fetching instructions using referring expressions. With the MTCM-AB being an extension of the MTCM \cite{magassouba2019understanding}, the multimodal language understanding for fetching instruction (MLU-FI) task context is similar to the one defined \cite{magassouba2019understanding}. Let us summarize and recall this task context below.

\subsection{Task Description}
Our aim is to predict a target referred by an initial instruction among a set of candidate targets in a visual scene. Instructions are not constrained which is more natural but increases the complexity of the comprehension task because users may use referring expressions to characterize a target. Fetching instructions based on referring expressions can be from the following types `Take the Kleenex box and put it in the bottom right box' or `Go to the kitchen and take the tea bottle on the upper shelf'.  
To address the MLU-FI, similar inputs and outputs, such as those in \cite{magassouba2019understanding} are considered:
\begin{itemize}
      \item[$\bullet$]{\bf Input}: A fetching instruction as a sentence in addition to an image of the scene.
      \item[$\bullet$]{\bf Output}: The most likely the target-source pair.  
\end{itemize}
The terms {\it target} and {\it source} are defined as follows.
 \begin{itemize}
     \item[$\bullet$]{\bf Target}: A daily object (e.g. bottle or snacks) that a user intends the robot to fetch 
     \item[$\bullet$]{\bf Source}: The origin of the target (e.g. desk or cabinet)
\end{itemize}
The evaluation metric is the prediction accuracy based on the top-1 target prediction. This metric is commonly used in the visual semantic embedding methods for the MLU-FI \cite{hatori2018interactively, Shridhar-RSS-18, magassouba2019understanding} and allows comparisons on standard datasets.
Additionally, we do not rely on dialogue systems to disambiguate the target from the candidate targets unlike \cite{hatori2018interactively} and \cite{Shridhar-RSS-18}. 
Ultimately this study does not focus on object detection. We suppose that the bounding boxes of the target and source are given in advance.

\begin{figure*}[t]
   \centering
      \includegraphics[scale=0.10]{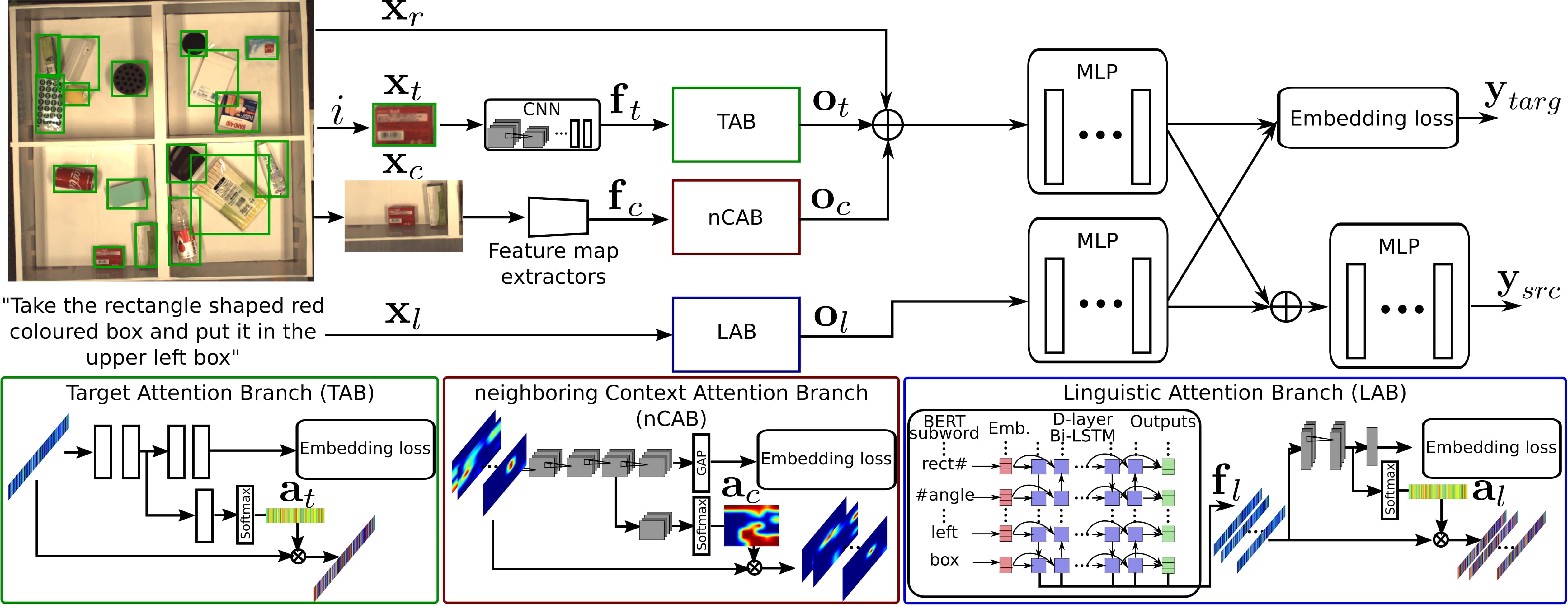}
      \caption{Proposed method framework:  the MTCM-AB utilizes three attention branches for the target (TAB), the visual context (nCAB) and the linguistic inputs. A Hinge loss embedding is used in these branches. In the main network the target and source are predicted from the attended features. }
   \label{fig:method}
 \squeezeup
\squeezeup
 \end{figure*}
 
\subsection{Task Background}
The MTCM-AB is not specifically designed for a given scene or context. Although this approach could be used for various scenarios, our method is validated on two types of dataset, in real and simulated environments described below.

\subsubsection{Home environment}
In this configuration, we use a simulation-based dataset from the  Partner Robot challenge-Virtual Space (WRS-PV) competition that uses SIGVerse \cite{inamura2013development}. A simulated environment allows repeated and varied tasks for a small cost compared with real environment, which makes this choice reasonable.
WRS-PV depicts home environments as represented in Fig. \ref{fig:sigverse}. The  three-dimensional environments (Unity-based) are augmented to make them more realistic. In this environment, a targeted DSR, that is HSR (Human Support Robot), is able to freely navigate and manipulate objects. In this context, the MTCM-AB predicts the most likely target among several candidates for instructions such as ``{\it Go to the bedroom to get me the pink toy on the wooden wagon}''.

\subsubsection{Pick-and-place scene}
Additionally a standard dataset PFN-PIC \cite{hatori2018interactively} for multimodal language understanding  is used. This dataset was designed for pick-and-place tasks from an armed robot with a top-view camera. The scene consists of four boxes, in which several candidate targets (up to $40$) are randomly placed. Each target is annotated with pick-and-place instructions such as ``{\it Move the red bottle near the coke can to the top left box}''.
Note that only the picking instruction comprehension is solved in this study.

\section{Proposed method}\label{sec:method}

\subsection{Attention Branch Network}
Our method consists of target prediction with respect to  an instruction in natural language.
We extend the MTCM \cite{magassouba2019understanding} with attention mechanisms that are used to improve the prediction from the linguistic and visual inputs. These attention mechanisms are inspired by the ABN, which was initially proposed in \cite{Fukui_2019_CVPR}.
In the ABN approach, the CAM is extended to produce an attention mask for improving image classification. However, instead of directly using the attention network into the classifier, the ABN is decomposed into parallel branches to avoid deteriorating the classifier accuracy. The branches refer to the following sub-components of the ABN: 
\begin{enumerate}
    \item an attention branch that produces attention maps.
    \item a prediction branch  that predicts the likelihood of some label.
\end{enumerate}
Both the attention and prediction branches of the ABN are classifiers. The attention maps are derived from the predicted label in the attention branch. Hence, this type of attention is built in a supervised manner. 
As an extension of CAM networks, the ABN also allows visual explanation when extracting the attention maps. Such a feature is particularly desirable for qualitatively validating this approach.
\Update In this study, we extend the MTCM to the MTCM-AB by introducing the ABN to multimodal language understanding as detailed below. Similar to the MTCM, the target and source are predicted from the full sentence. The training procedure is outlined in Algorithm 1.
\Done

\subsection{Novelty}
We propose the MTCM-AB (see Fig. \ref{fig:method}) which can focus on the relevant expressions in a sentence and their visual representation by using attention branch mechanisms. The MTCM-AB has the following characteristics.
\begin{itemize}
\item The MTCM-AB extends the MTCM, with attention branches to focus on the salient part of linguistic and visual inputs for referring expression comprehension. 

\item We introduce several attention branches for linguistic and visual features.  A neighboring context branch allows the network to predict a given target candidate based on the neighbor landmarks.

\item A visual explanation is produced from the attention maps generated by the MTCM-AB.
\end{itemize}

\subsection{MTCM-AB}
\subsubsection{Inputs}
The  MTCM-AB takes as input visual and linguistic features in a similar manner to most visual semantic embedding methods.
We assume that for each target candidate $i \in \{1, ..., N\}$ and source $i^{'} \in \{1,...,M\}$, their respective cropped image and positions are made available. Given a target candidate, the set of inputs ${\bf x} (i)$ is defined as:
\begin{equation}\label{equ:x}
    {\bf x}(i)=\{{\bf x}_{l}(i), {\bf x}_{t} (i), {\bf x}_{c} (i),  {\bf x}_{r} (i) \},
\end{equation}
where ${\bf x}_{l}(i)$, ${\bf x}_{t}(i)$, ${\bf x}_{c} (i)$ and ${\bf x}_{r}(i)$ denote linguistic, target, context and relation features. We purposefully omit index $i$ in the following,  that is,  ${\bf x} (i)$ is then written as ${\bf x}$,  when further clarity is not necessary.

Visual input ${\bf x}_t$ is defined as the cropped image of the target, and ${\bf x}_{c}$ is a cropped image that characterizes a target and its neighborhood. The latter input ${\bf x}_{c}$ is more thoroughly defined in the next section. Linguistic input ${\bf x}_l$ consists of sub-word vector embedding whereas input ${\bf x}_{r}$ is a vector characterizing the position of the target candidate in the environment (e.g., other target candidates, location in the scene, location with respect to the source).

\subsubsection{Linguistic Attention Branch}
The purpose of the linguistic attention branch (LAB) is to emphasize the most salient part of the linguistic features for instruction comprehension. Similar to our previous approach \cite{magassouba2019understanding}, BERT \cite{devlin2018bert} is used in the MTCM-AB for sub-word embedding. The BERT model uses Transformers \cite{vaswani2017attention} and a sub-word masking system for language embedding. In \cite{magassouba2019understanding}, BERT is reported to be better than simple embedding vectors.
From the sub-word embedding, a multi-layer Bi-LSTM network is used to  obtain a latent space representation of the linguistic features. The last hidden states of each layer are concatenated to form linguistic feature maps ${\bf f}_l$  from which a linguistic attention mask is extracted using the same principle as the ABN. 
Feature maps ${\bf f}_l$ are processed through one-dimensional convolutional layers followed by a single fully connected layer (FC). Linguistic attention map ${\bf a}_l$ is obtained from the second convolutional layer  that is convoluted with an additional layer and normalized by a sigmoid activation function. This attention map selectively focuses on an area of the LSTM final state that also encodes all the previous states. 
The output visual feature maps are then obtained using a masking process given by:
\begin{equation}\label{equ:ling_att}
    {\bf o}_{l}= {\bf a}_l \odot {\bf f}_{l},
\end{equation}
where $\odot$ denotes the Hadamard product. 
The LAB optimizes an embedding loss $J_{l}$ defined in the Section \ref{sub:loss}.

\subsubsection{Target Attention Branch}  
The target attention branch (TAB) produces an attention map for the candidate target images. In this branch, the input ${\bf x}_t$ is transformed into a latent space feature ${\bf f}_t$ through a CNN.  Feature ${\bf f}_t$ is  processed into FC layers. Similarly to the LAB, a visual attention map  ${\bf a}_t$ is extracted from the second FC layer that is processed in a parallel branch composed of a FC layer and a sigmoid activation function. Output latent space feature ${\bf o}_t$ is then obtained by 
\begin{equation}\label{equ:vis_att}
    {\bf o}_{t}= {\bf a}_t \odot {\bf f}_{t}.
\end{equation}
The  TAB also optimizes a loss function $J_{t}$ to predict the attention mask.

\subsubsection{neighboring Context Attention Branch}
One of the main novelties of the MTCM-AB is the neighboring context attention branch (nCAB). Instead of analytically defining context features, we use an attention branch mechanism to focus on the relevant part of the image in the surroundings of a given target. For instance, in the instruction ``{\it Take the apple near that is near the teddy bear}'', the teddy bear is a crucial landmark that should be used to predict the correct target, especially when there are several apples in the scene. The nCAB allows us to capture such relationships.  In \cite{yu2017joint} these relationships were defined analytically and only with the object of the same class (e.g, apple only), whereas the MTCM-AB uses an attention branch.

Thus, an extended cropped image of the candidate target is extracted from the image. This extended cropped image is denoted as ${\bf x}_{c}$. This input is encoded into feature maps  ${\bf f}_{c}$ from a CNN feature extractor. Inspired by the CAM structure \cite{zhou2016cvpr}, visual feature maps ${\bf f}_{c}$ are processed with four two-dimensional convolutional layers. These convolutional layers are followed by a global average pooling (GAP). In parallel, context attention map ${\bf a}_{c}$ is created from an additional convolution and sigmoid normalization of the third convolutional layer. The output context feature maps are given by
\begin{equation}\label{equ:rv_att}
    {\bf o}_{c}= {\bf a}_{c} \odot {\bf f}_{c}.
\end{equation}
The loss function of the nCAB is noted $J_{c}$.
\subsubsection{Perception Branch}
The perception branch follows a classic visual semantic embedding structure.  Indeed, the visual linguistic and relation features are encoded to share a common latent space. A visual multi-layer perceptron (MLP) encodes the concatenation of ${\bf o}_{t}$, ${\bf o}_{v}$ and ${\bf x}_{r}$. In parallel a linguistic MLP encodes linguistic features ${\bf o}_{l}$.  Both MLP ouputs are used to compute the embedding loss defined in the following section.
The source is predicted as  ${\bf y}_{src}$ from a third MLP that combines the two previous MLP outputs. 
To predict the correct source, a cross-entropy loss $J_{src}$ is used:
\begin{align} \label{equ:J_src}
    J_{src} &= -\sum_n \sum_{m} y^{*}_{nm} \log p(y_{nm}),
\end{align}
where $y^{*}_{nm}$ denotes the label given to the $m$-th dimension of the $n$-th sample, and $y_{nm}$ denotes its prediction.

\subsubsection{Loss functions}\label{sub:loss}
The MTCM-AB is trained by minimizing several embedding loss functions related to the different branches. All of them are based on a Hinge loss model. This loss function consists in increasing the similarity between appropriate pairs of linguistic and non-linguistic features  while decreasing the similarity  between  inappropriate  ones. The network then minimizes the global loss function $J_{total} =  \lambda_c J_{c} +  \lambda_t J_{t}  + \lambda_l J_{l} + \lambda_p J_{p} +  \lambda_{src} J_{src}$. The parameters $\lambda_i$ are loss weights that are defined in the experimental section.
In the perception branch, loss function $J_{p}$ is expressed as a triplet Hinge loss

\begin{equation}
\begin{split}\label{equ:J}
J_p &=\sum_i \bigl\{
   	\max \bigl(0, \lambda_M + f(g_1(i),g_2(j) )
   	- f(g_1(i),g_2(i) \bigr) \\
   	&+ \max \bigl( 0, \lambda_M + f(g_1(k), g_2(i) )
   	- f(g_1(i),g_2(i) \bigr) \bigr\},
\end{split}
\raisetag{2\normalbaselineskip}
\end{equation}
where $\lambda_M$ is the margin, and $f(\cdot,\cdot)$ is a similarity function (e.g., cosine similarity). Functions $g_1(\cdot)$ and $g_2(\cdot)$ are the networks related to the linguistic and non-linguistic features, respectively. The incorrect linguistic and non-linguistic features are extracted from two random candidate targets $j$ and $k$ in the same image as the current target $i$. Analytically,  $j$ and $k$ are sampled from $\{1,...,T\}$, where $j \not= i$ and $k \not= i$.

Given $J$ a generic notation of $J_{l}$, $J_{t}$ and $J_{c}$, their respective Hinge loss function is characterized by:
\begin{equation}\label{equ:J_p}
J =\sum_i \max \bigl(0, \lambda_M + f(g_1(i), g_2(j) )
   	- f(g_1(i), g_2(i) \bigr)
\end{equation}
\squeezeup
\squeezeup
\squeezeup

\Update \begin{algorithm}[h]
\SetAlgoLined
\KwInput{$\{{\bf x}_{l}, {\bf x}_{t}, {\bf x}_{c},  {\bf x}_{r} \}$}
\KwOutput{${\bf y}_{targ}$, ${\bf y}_{src}$}
\ForEach{\text{epoch}}{
  \with ($i \neq k \neq j$  and  $i$, $j$, $k$ in the same image:){
    Sample  $i$ and $j$ sentences , $i$ and $k$ vis. features\;
    }
  Obtain $({\bf o}_l, {\bf o}_t, {\bf o}_c)$ by Eq. (2)-(4)\;
  Predict $({\bf y}_{targ}, {\bf y}_{src})$\;
  Obtain $J_{total}$ by Eq. (6)-(7)\;
  Update network parameters\;
 }
 \caption{MTCM-AB training procedure} \label{alg:train}
\end{algorithm}
\Done
\squeezeup
\squeezeup
\squeezeup

\section{Experiments}\label{sec:exp}

\subsection{Dataset}
\subsubsection{The PFN-PIC dataset}

\begin{table}[b]
\squeezeup
\squeezeup
\small
\centering
\caption{\small Parameter settings and structures of MCTM-AB}\label{tab:param}
\begin{tabular}{|c|l|}

\hline
MTCM-AB & Adam (Learning rate= $2e^{-4}$, \\
 Opt. method  &$\beta_1=0.99$, $\beta_2=0.9$) \\
\hline
Bi-LSTM   &$3$ layers, $1024$-cell\\
\hline
 & MLP: $[1024]$ $\times$3 layers \\
 \cline{2-2}
 & TAB: $[2048]$ $\times$5 layers \\
\cline{2-2}
Num. & nCAB: $[10\times10\times512]$ $\times$3 layers \\
nodes & and   $[10\times10\times1]$  (attention) \\
\cline{2-2}
 & LAB: $[1\times1024\times6]$ $\times$4 layers\\ 
 & $[1\times1024\times1]$ (attention)\\
\hline
Weight   &$\lambda_c=1, \lambda_t=1$,  $\lambda_l=1$,  $\lambda_p=1$,  $\lambda_s=0.1$  \\
\hline
 Batch size & 128  \\
 \hline
 $\delta_c$ & [0, 25, 50, 75, 100, 125, 150]\\
\hline
\end{tabular}
\end{table}

In this experiment we evaluated our approach on the PFN-PIC dataset \cite{hatori2018interactively}, which allowed us to compare the MTCM-AB to other  proposed methods ({e.g.}, \cite{magassouba2019understanding}). PFN-PIC contains $89,891$ sentences in the training set and $898$ sentences in the validation set to instruct $25,861$ targets in the training set and 532 targets in the validation one (see Fig. \ref{fig:sigverse}). The sentences, on average 14.3 words, were given by three different annotators.

\subsubsection{The WRS-PV dataset}
In the second phase, we evaluated the MTCM-AB on a simulation-based dataset: WRS-PV (see Fig. \ref{fig:sigverse}). We used the same dataset collected in \cite{magassouba2019understanding} with additional instructions. 
The dataset is composed of $308$ images from which $2015$ instructions in the training set and 74 instructions in the validation set are provided. The dataset was labelled by two annotators. This dataset has an average of $3.4$ targets per image, and $10.7$ words for each instruction.

\subsection{Experimental Setup}
The experimental setup is summarized in Table \ref{tab:param}. The instructions were pre-processed and tokenized using the 24-layer pre-trained BERT model. Each sub-word, obtained by a word-piece model, were embedded into a 1024-dimensional vector. 
The feature extractor and CNN as shown in Fig. \ref{fig:method} were both based on ResNet\cite{he2016deep}. The input images were downscaled to $299 \times 299$ before being processed in ResNet-50. Feature maps ${\bf f}_{c}$ were extracted inside the 4$^{th}$ block of ResNet, while features ${\bf f}_{t}$ were obtained from the last convolutional layer of the network. For each MLP, we applied batch normalization and a ReLU activation function for three layers, except for the source prediction,  for which a softmax function was used on the last layer.
In the LAB, we used a three-layer Bi-LSTM with 1024-size cells. Feature maps ${\bf f}_l$ were processed in three convolutional layers. In the TAB, four fully-connected layers were used, whereas the nCAB was based on four convolution layers.
The different loss weights were all set to one except the source loss set to $0.1$. 
We also introduced a parameter $\delta_c$ that represents the variation in size of the context input ${\bf x}_c$. Indeed, the size of ${\bf x}_c$ corresponds to the size of the cropped image target to which is added $\delta_c$ in width and height. \Update Considering the size of the input image, we selected $\delta_c = [0, 25, 50, 75, 100, 125, 150]$ pixels during the experiments. \Done

The MTCM-AB had  27.5 M parameters and was trained on a RTX 2080Ti with 12 GB of GPU memory, 64 GB RAM and a Intel Core i9 3.6 GHz processor. The results were reported after 100 epochs when the training loss was reduced by  approx. 95\%. With this setup, it took approx. five hours to train the MTCM-AB on average for the PFN-PIC dataset. This time was decreased to approx. three hours and a half by using a bigger batch size with two Tesla V100 GPUs.

The same parameters as in  Table \ref{tab:param} were used to evaluate WRS-PV dataset, except the learning rate that was decreased to 5$\times$10$^{-5}$ and a batch size of 64.
\subsection{Quantitative Results}

\Update    
\begin{table}[t]
\normalsize
\caption{\Update \small Top-1 accuracy on PFN-PIC and WRS datasets \Done}
\label{tab:results}
\centering
\begin{tabular}{l|c|c}
\hline
&\multicolumn{2}{c}{\bf Top-1 Target accuracy $[\%]$} \\
\cline{2-3}
{\bf Method }&\multicolumn{1}{c|}{ \bf PFN-PIC} &\multicolumn{1}{c}{ \bf WRS-PV}  \\
\hline
\hline
Hatori et al. \cite{hatori2018interactively} & 88.0 & $-$ \\
\hline
MTCM \cite{magassouba2019understanding}& 88.8 $\pm$ 0.43 & 87.0  $\pm$ 1.81  \\
\hline
MTCM-AB ($\delta_c=50$) & {\bf 90.1 $\pm$ 0.47}  & {\bf 89.2  $\pm$ 1.33}\\
\hline
Human Performance & 90.3  $\pm$ 2.01  &  94.3 $\pm$ 3.98   \\
\hline
\end{tabular}
\bigskip

\normalsize
\caption{\Update \small MTCM-AB accuracy with respect to $\delta_c$  \Done}
\label{tab:delta}
\centering
\begin{tabular}{l|c|c}
\hline
\multicolumn{1}{l|}{\bf Extension}&\multicolumn{2}{c}{\bf Top-1 Target accuracy $[\%]$} \\
\cline{2-3}
\multicolumn{1}{l|}{$\boldsymbol{\delta}_c$ }&\multicolumn{1}{c|}{ \bf PFN-PIC} &\multicolumn{1}{c}{ \bf WRS-PV}  \\
\hline
\hline
0 & 89.8 $\pm$ 0.22 & 88.7 $\pm$ 1.81 \\
\hline
25& 89.3 $\pm$ 0.39 & 89.4 $\pm$ 1.76  \\
\hline
50 & {\bf 90.1 $\pm$ 0.47 }& 89.2 $\pm$ 1.33 \\
\hline
75 & 90.0 $\pm$ 0.33 & {\bf 90.8 $\pm$ 1.22} \\
\hline
100& 90.0 $\pm$ 0.26  & 89.5 $\pm$ 1.71 \\
\hline
125 & 89.7 $\pm$ 0.21  &  89.5 $\pm$ 1.48   \\
\hline
150 & 89.4 $\pm$ 0.29  & 89.2 $\pm$ 1.35   \\
\hline
\end{tabular}
\squeezeup
\squeezeup
\end{table}
\Done

\begin{figure*}[t]
  \subfloat[]{\includegraphics[width=4.3 cm, height=3.5cm]{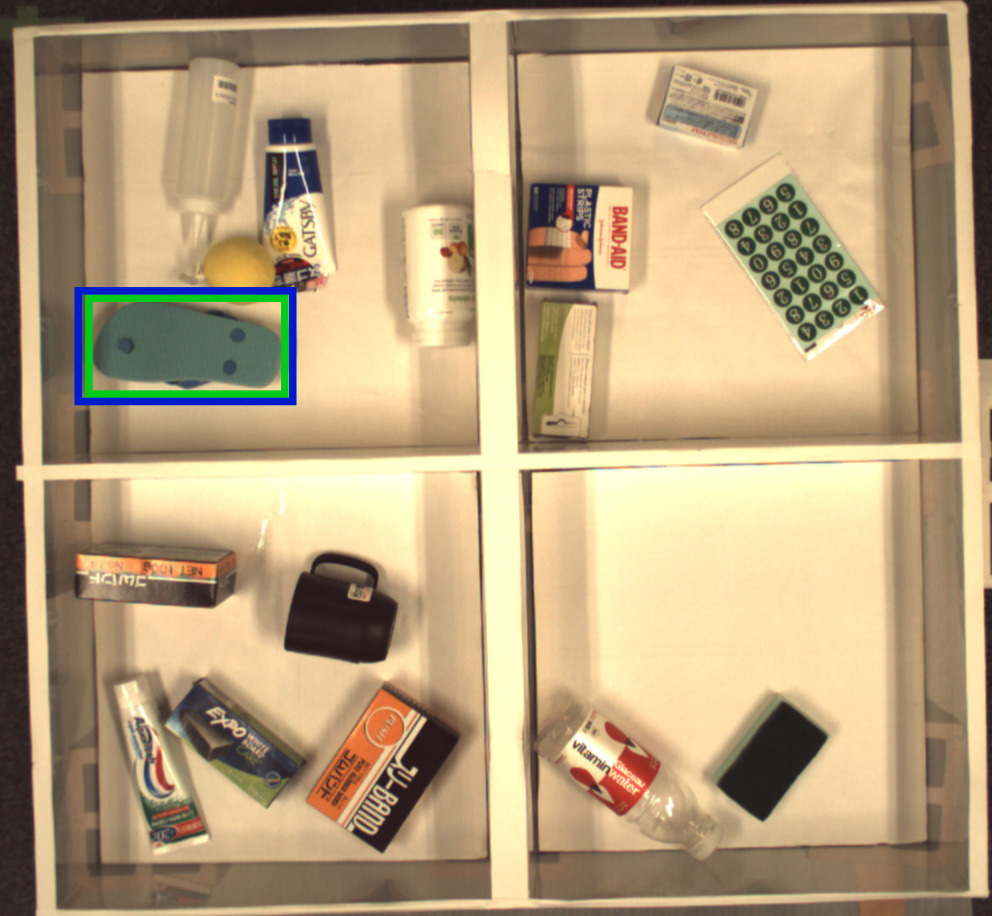}}\enskip
  \subfloat[]{\includegraphics[width=4.3 cm, height=3.5cm]{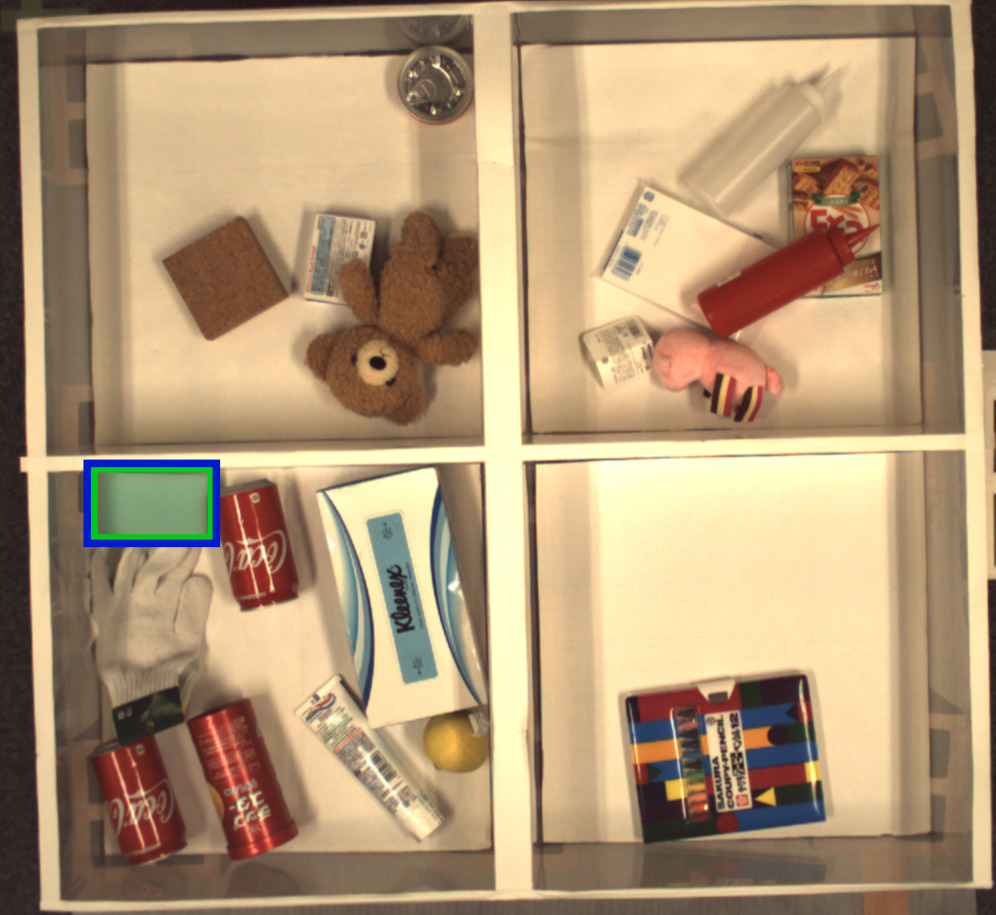}}\enskip
  \subfloat[]{\includegraphics[width=4.3 cm, height=3.5cm]{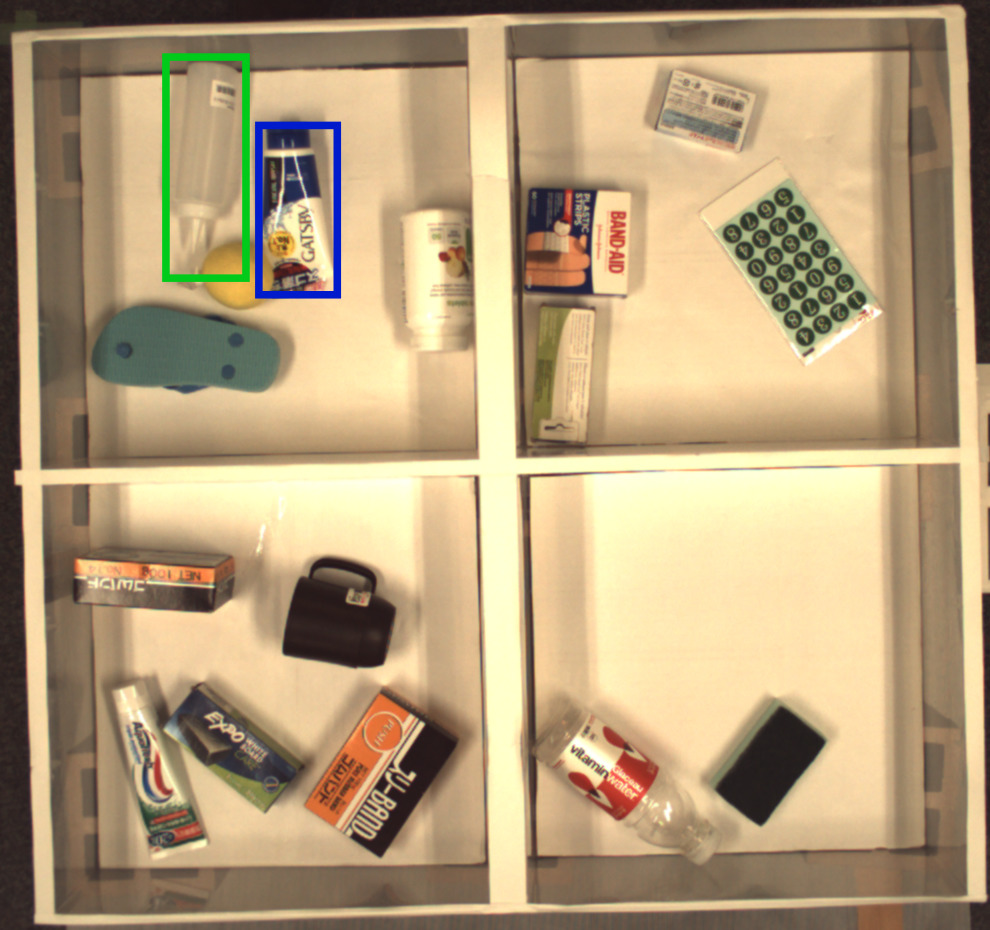}}\enskip
  \subfloat[]{\includegraphics[width=4.3 cm, height=3.5cm]{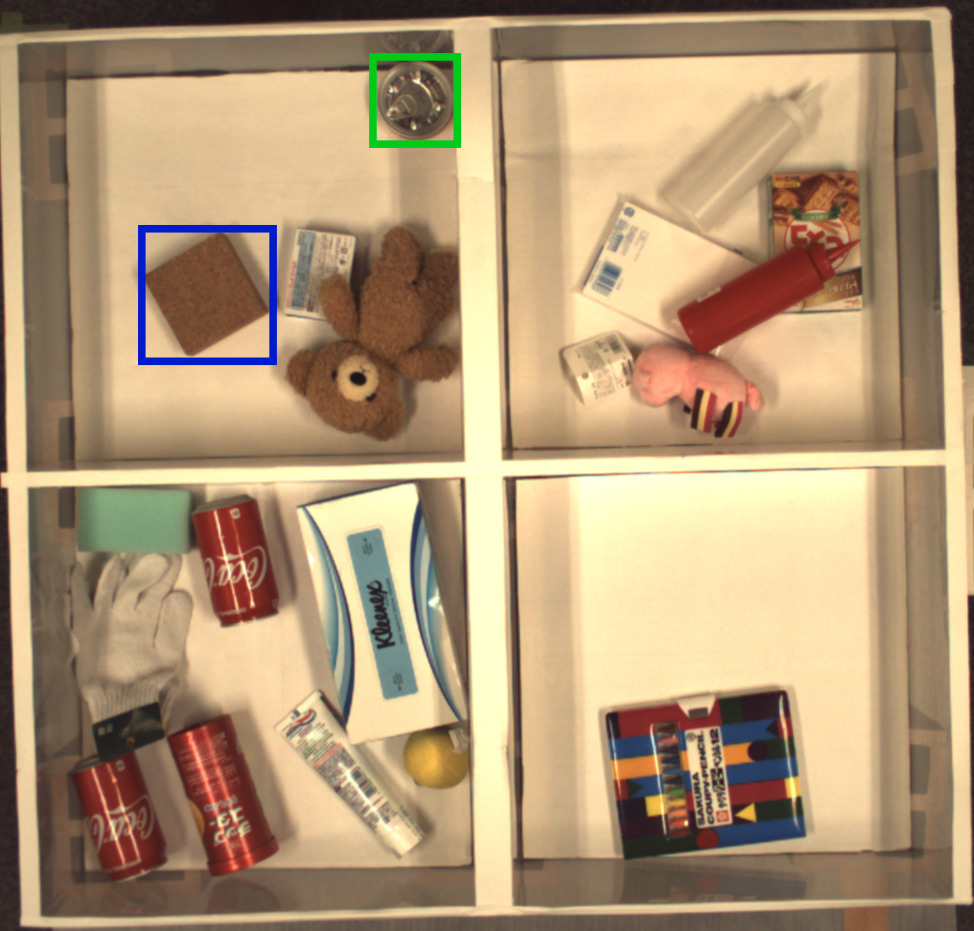}}\\ \squeezeup
  
  \subfloat{\includegraphics[width=4.3 cm, height=3.5cm]{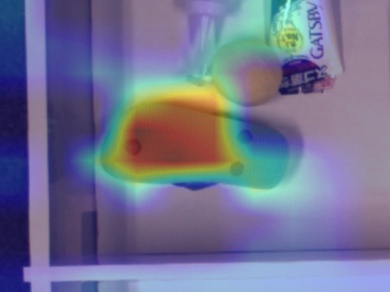}}\enskip
  \subfloat{\includegraphics[width=4.3 cm, height=3.5cm]{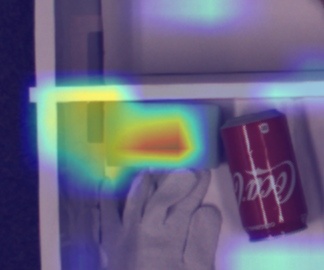}}\enskip
  \subfloat{\includegraphics[width=4.3 cm, height=3.5cm]{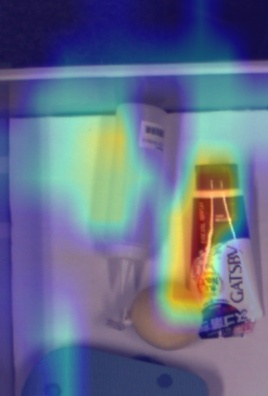}}\enskip
  \subfloat{\includegraphics[width=4.3 cm, height=3.5cm]{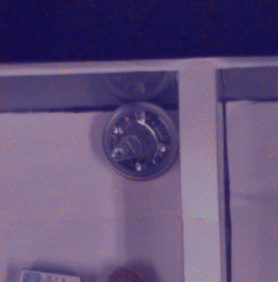}}
    \caption{\small Qualitative results of the MTCM-AB.  In the first row the prediction is given in blue while the ground truth is in green. The attended region of each context feature ${\bf x}_c$  is given in the second row. The two first columns refer to correct predictions. The third column refers to an erroneous prediction ({\it i.e.} wrongly attended target), while the last column refers to an erroneous prediction due to an incorrect ground truth (``brown pack'' is instructed but ``can'' is given the label). \\
    (a): ``Take the blue sandal move it to lower left box''. (b): ``Take the green item next to the pair of white gloves and move it to top left box''. (c): ``Move the grey colored bottle at top left to box below''. (d): ``Pick the brown pack and put it in lower left box''. }
    \label{fig:samples}
    \squeezeup
    \squeezeup
\end{figure*}

The quantitative results are presented in Table \ref{tab:results}. The \mbox{top-1} accuracy is the evaluation metric as defined in \cite{yu2017joint}. This evaluation metric correspond of the accuracy obtained for the most likely target ({e.g.}, highest cosine similarity) predicted, given an instruction.
Quantitative results of the baseline methods \cite{hatori2018interactively} and \cite{magassouba2019understanding} are also reported for comparison. Except for \cite{hatori2018interactively} method, the average accuracy and standard deviation is provided for five trials.  \Update In the case of the MTCM-AB, we report additional results in Table \ref{tab:delta} with varying sizes of ${\bf x}_{c}$ by setting $\delta_c$ with a 0 to 150-pixel wise extension.
On PFN-PIC,  the MTCM-AB outperformed  the MTCM and baseline method  \cite{hatori2018interactively} by 1.3\% and 2.1\% respectively. The results of WRS-PV corroborated the trend observed on the PFN-PIC dataset. The MTCM-AB attention mechanism improved the target prediction  accuracy by 2.5\% on average.\Done

Furthermore, we conducted an evaluation on five test subjects. This evaluation was performed by selecting randomly $200$ samples of the validation set of PFN-PIC and the full validation set of WRS-PV. For each sample, each test subject was given the instruction, and had to select the most likely target among a set of candidate targets in the same image. The accuracy  of these experiments are reported in Table \ref{tab:results} through the human performance row. 
From these evaluations,  it appears that the human-level comprehension of the PFN-PIC dataset, that can be considered as an upper-bound, is on average at 90.3\% of accuracy with 2.01\% of standard deviation.  Thus, on the PFN-PIC dataset there is no statistically significant difference between human performance and the MTCM-AB. In the case of the WRS-PV dataset, the human performance was 94.3 \% on average with 3.98\% of standard deviation. Although test subjects performed with higher accuracy, the difference in performance with the MTCM-AB remains reasonable considering the dataset size.

\Update As reported in Table \ref{tab:delta}, from $\delta_c$ analysis, we found that increasing the size of the neighboring context from 50 to 150 pixels did not improve the accuracy for the PFN-PIC dataset. Although it may be thought that by increasing the size of $\delta_c$ more information about the context can be captured, we do believe that it also increases the probability of attending to an erroneous region. Indeed, the PFN-PIC dataset is highly cluttered with relatively small objects, which require accurate attention maps. This is exemplified by $\delta_c=[125, 150]$ which has a lower accuracy than the configuration with $\delta_c=0$. Regarding the WRS-PV dataset, the optimal value was found at 75 pixels. As the dataset is less cluttered,  setting $\delta_c>0$ always improves the accuracy.

\Done

To characterize the contribution of each attention branch, we also report the results of an ablation study in Table \ref{tab:ablation} for the PFN-PIC dataset.  These results clearly support the contribution of the attention branches and the novelty of our approach. Both linguistic and visual attention branches improved the prediction accuracy compared to the MTCM. 

\begin{table}[b]
\squeezeup
\normalsize
\caption{\small Ablation study of the MTCM-AB for PFN-PIC dataset}
\label{tab:ablation}
\centering
\begin{tabular}{l|c}
\hline
{\bf Method }& \multicolumn{1}{c}{\bf Top-1  $[\%]$} \\
\hline
\hline
MTCM & 88.8 $\pm$ 0.43\\
\hline
MTCM-AB (LAB only) & 89.2 $\pm$ 0.43 \\
\hline
MTCM-AB (LAB + TAB) & 89.6 $\pm$ 0.28 \\
\hline
MTCM-AB (LAB + TAB + nCAB) & {\bf 90.1 $\pm$ 0.47}\\
\hline
\end{tabular}
\squeezeup
\squeezeup
\end{table}

\begin{figure*}[tp]
  \centering
   \subfloat[]{\includegraphics[width=4.3 cm, height=3.5cm]{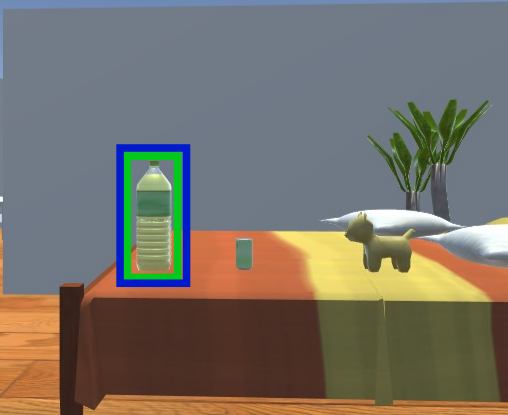}}\enskip
    \subfloat[]{\includegraphics[width=4.3 cm, height=3.5cm]{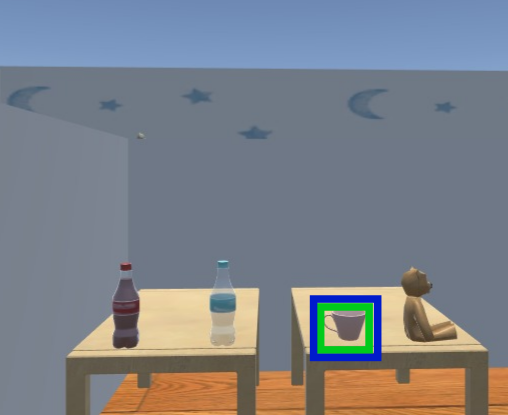}}\enskip
     \subfloat[]{\includegraphics[width=4.3 cm, height=3.5cm]{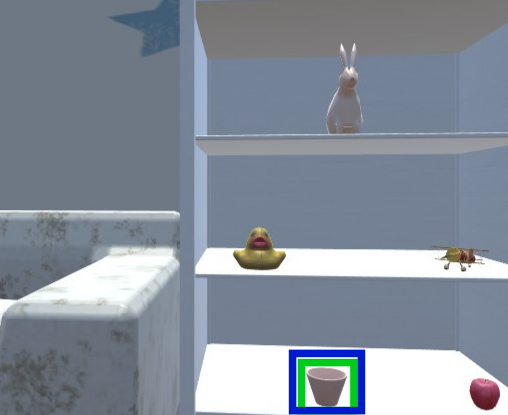}}\enskip
      \subfloat[]{\includegraphics[width=4.3 cm, height=3.5cm]{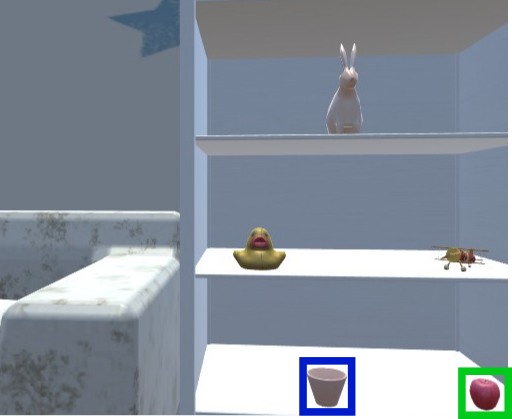}}\enskip\\
     \squeezeup
    \subfloat{\includegraphics[width=4.3 cm, height=3.5cm]{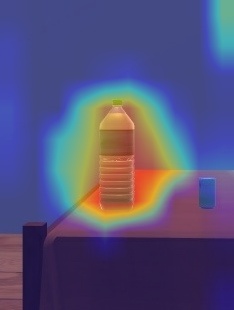}}\enskip
    \subfloat{\includegraphics[width=4.3 cm, height=3.5cm]{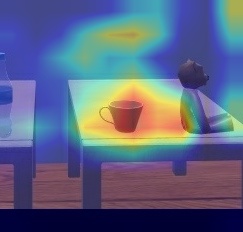}}\enskip
    \subfloat{\includegraphics[width=4.3 cm, height=3.5cm]{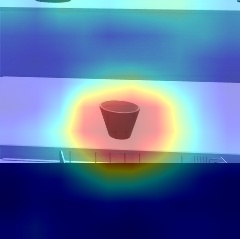}}\enskip
    \subfloat{\includegraphics[width=4.3 cm, height=3.5cm]{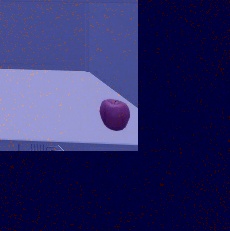}}\enskip
    \caption{\small  Qualitative results on WRS-PV dataset. The predicted target is reported in blue while the ground truth is in green. The attention maps are reported on the second row. \\
    (a)  Bring me the yellow bottle from the bed with orange sheets. (b) Take a coffee cup on the right-hand table. (c) Give me the pink cup on the lower row of the shelf (d) Take an apple on the same shelf that a coffee cup is placed.}  \label{fig:wrs_quali} 
 \squeezeup
  \squeezeup
\end{figure*}

\subsection{Qualitative Results}
\subsubsection{PFN-PIC qualitative results}
The qualitative results shown in Fig. \ref{fig:samples} illustrate predictions for  given fetching instructions. Each column  of the first row represents the MTCM-AB results for different scenarios. The second row illustrates the attention map of each sample given the context feature. Subfigures (a) and (b) show correct predictions. The attended regions show that the MTCM-AB focuses on the instructed target. The third column illustrates an erroneous case. Here the attention focused on the incorrect candidate target leading to a wrong prediction. The last column is also an erroneous prediction.  However in this case, the ground truth label is incorrect. The target is not a ``brown pack'' as specified in the instruction, but a ``can''. Therefore, it is reasonable that the MTCM-AB did not attend the ``can''.

\subsubsection{ WRS-PV qualitative results}
Qualitative results on the WRS-PV dataset (see Fig. \ref{fig:wrs_quali}) can also be analyzed in the same way. The three first samples illustrate correct prediction with consistent attended regions. The last sample, with the instruction ``Take an apple on the same shelf that a coffee cup is placed'', is erroneous and our model predicts the cup instead of the apple. The attention map shows that the apple is not attended unlike the cup. This kind of mistake may be caused by the linguistic understanding since both the apple and cup are mentioned in the instruction.

\subsection{Error Analysis} 
A thorough analysis of the MTCM-AB results allows us to characterize the different failure cases of our approach for the PFN-PIC dataset. We categorize the erroneous predictions based on main cause as follows:
\begin{itemize}
 \item ES (erroneous sentence): the ground truth does not correspond to the target specified in the instruction (see Fig. \ref{fig:samples}-d)
 \item NE (negation): the ground truth is specified from a negation sentence which is thought to be difficult to solve in NLP community; {e.g.}, ``grab the glue not yellow and move...'' and the predicted target is the yellow glue. In such a situation, the linguistic encoding should  first be able to extract the negation features. Then the network should interpret `not yellow'' that covers a wide range of different color characteristics (green, blue,...).
\item REL (relation to landmark): the ground truth is defined with respect to landmark objects and the predicted target position is incorrect with respect to this landmark; {e.g.}, ``move the rectangular card next to the red ketchup bottle...''  and the predicted target is far from the ketchup.
\item RES (relation to source):  the ground truth target is defined with respect to a source and the predicted target position is incorrect with respect to this source; {e.g.}, ``grab the black object in the middle right of the bottom left box...'' and the predicted target is in the upper left corner of the bottom left box.
\item SE (source error): the instruction specifies a given source and the predicted target position is in a different source; {e.g.}, ``grab small white plastic container on lower right box...'' and the predicted target in on the upper right box. 
\item SU (sentence understanding): the predicted target characteristics are different from the specified ground truth characteristics; {e.g.}, ``grab thin orange and black box...'' and the predicted target is a large orange and black box.
\item TR (target recognition): the predicted target is visually similar to the ground truth target but incorrect;
{e.g.}, in Fig. \ref{fig:samples}-b, a sample predicted the red tube instead of the coke can that looks very similar.
\item O (others):  this category  includes a case where the instruction contains words that rarely appear in the training samples; {e.g.},  the word ``barrel'' is in the instruction while it appears only once in the training set. A meaningful encoding of this word is complex for the network.

\end{itemize}
The results are reported on Table \ref{tab:err}. Despite being superior to the MTCM, the  major errors of the MTCM-AB are related to referring expression based on landmarks, sentence understanding and wrongly recognized targets. In particular, the parameter $\delta_c$ may affect the  error rate of REL and need a careful tuning. Some improvement in the linguistic encoding may also be envisioned to increase the accuracy. Drawing inspiration from the recent methods for natural language processing (NLP), transformer-based model for language understanding such as XLNet \cite{yang2019xlnet} represent a path of improvement of our current Bi-LSTM model. Likewise, the TR error rate may be decreased with more sophisticated image classifier than ResNet-50.

\Update
\begin{table}[t]
\normalsize
\caption{\small \Update Categorization of the erroneous predictions of the MTCM-AB on the PFN-PIC dataset \Done}
\label{tab:err}
\centering
\begin{tabular}{l|l|rr}
\hline
{\bf Error ID }&{\bf Description }& \multicolumn{2}{c}{\bf  Error $\#$ } \\
\hline
\hline
ES &Erroneous sentence & 18 &(21 \%)\\
\hline
NE & Negation  & 5 &(6 \%)\\
\hline
REL & Relation to landmark & 20 &(24 \%)\\
\hline
RES & Relation to source   & 4 &(5 \%)\\
\hline
SE & Source error  & 8 &(9 \%) \\
\hline
SU & Sentence understanding  & 12 &(14 \%) \\
\hline
TR & Target recognition  & 16 &(19 \%)\\
\hline
O & Others  &2 &(2 \%)\\
\hline
\hline
\multicolumn{1}{l|}{\bf Total} &$-$ & 85 & (100 \%) \\
\hline
\end{tabular}
\squeezeup
\squeezeup
\end{table}
\Done

\section{Conclusion}
In a context of high demand for responsive domestic service robots, we proposed the MTCM-AB which predicts the likelihood of targets to fetch given an instruction and a visual input. The following contributions of this study can be emphasized:
\begin{itemize}
 \item The MTCM-AB extends the MTCM with attention branches for linguistic and visual features. Our method achieves higher accuracy than the MTCM. Actually, our method is close to a human-level accuracy on a standard dataset.
\item  We showed that multimodal attention achieves higher accuracy than monomodal attention on linguistic or visual inputs.
\item We qualitatively validated the MTCM-AB performance by showing that the instructed target was attended in the visual scene.
\end{itemize}
In future work, we plan to address the multimodal language understanding with only dense features (i.e., no heuristic inputs such as relation features) by improving the attention mechanism. \Update Similarly, we intend to develop a word-level attention mechanism to visualize the effect of linguistic attention on ambiguous instructions.\Done


\section*{Acknowledgements}
\vspace{-1mm}
This work was partially supported by JST CREST, SCOPE and NEDO.
\vspace{-3mm}

\bibliographystyle{IEEEtran}
\bibliography{strings,bib/bibthese}

\end{document}